\documentclass[sigconf]{acmart}

\usepackage{graphicx}
\usepackage{graphics}
\usepackage{tabularx}
\usepackage{multirow}
\usepackage{subfigure}
\usepackage{balance}

\usepackage[misc]{ifsym}

\AtBeginDocument{%
  \providecommand\BibTeX{{%
    \normalfont B\kern-0.5em{\scshape i\kern-0.25em b}\kern-0.8em\TeX}}}

\setcopyright{acmcopyright}
\copyrightyear{2021}
\acmYear{2021}
\acmDOI{10.1145/1122445.1122456}

\acmConference[Chengdu '21]{Chengdu '21: ACM Multimedia 2021}{Octoboer 21--24, 2021}{Chengdu, China}
\acmBooktitle{Chengdu '21: ACM Multimedia,
  Octoboer 21--24, 2021, Chengdu, China}
\acmPrice{15.00}
\acmISBN{978-1-4503-XXXX-X/18/06}

\begin{document}

\title{Text is NOT Enough: Integrating Visual Impressions into Open-domain Dialogue Generation}


\author{Lei Shen$^{1,2}$,\quad Haolan Zhan$^2$,\quad Xin Shen$^3$,\quad Yonghao Song$^{1}$,\quad Xiaofang Zhao$^{1}$*}


\makeatletter
\def\authornotetext#1{
\if@ACM@anonymous\else
    \g@addto@macro\@authornotes{
    \stepcounter{footnote}\footnotetext{#1}}
\fi}
\makeatother
\authornotetext{Corresponding author.}

\affiliation{
 \institution{\textsuperscript{\rm 1}Institute of Computing Technology, Chinese Academy of Sciences, Beijing, China}
 \institution{\textsuperscript{\rm 2}University of Chinese Academy of Sciences, Beijing, China \qquad \textsuperscript{\rm 3}Australian National University \country{Australia}}
 }
\email{shenlei17z@ict.ac.cn, zhanhaolan316@gmail.com, u6498962@anu.edu.au, {songyonghao, zhaoxf}@ict.ac.cn}

\def\authors{Lei Shen, Haolan Zhan, Xin Shen, Yonghao Song, Xiaofang Zhao}

\renewcommand{\shortauthors}{Shen and Zhan et al.}

\begin{abstract}





Open-domain dialogue generation in natural language processing (NLP) is by default a pure-language task, which aims to satisfy human need for daily communication on open-ended topics by producing related and informative responses.
In this paper, we point out that hidden images, named as \textit{visual impressions} (VIs), can be explored from the text-only data to enhance dialogue understanding and help generate better responses. Besides, the semantic dependency between an dialogue post and its response is complicated, e.g., few word alignments and some topic transitions. Therefore, the visual impressions of them are not shared, and it is more reasonable to integrate the response visual impressions (RVIs) into the decoder, rather than the post visual impressions (PVIs). However, both the response and its RVIs are not given directly in the test process.
To handle the above issues, we propose a framework to explicitly construct VIs based on pure-language dialogue datasets and utilize them for better dialogue understanding and generation. Specifically, we obtain a group of images (PVIs) for each post based on a pre-trained word-image mapping model. These PVIs are used in a co-attention encoder to get a post representation with both visual and textual information. Since the RVIs are not provided directly during testing, we design a cascade decoder that consists of two sub-decoders. The first sub-decoder predicts the content words in response, and applies the word-image mapping model to get those RVIs. Then, the second sub-decoder generates the response based on the post and RVIs.
Experimental results on two open-domain dialogue datasets show that our proposed approach achieves superior
performance over competitive baselines.
\end{abstract}




\maketitle

\section{Introduction}
\label{sec:intro}


Building intelligent open-domain dialogue agents that can converse with humans smoothly and establish long-term connections with users is a challenging task of artificial intelligence (AI) \cite{huang2020challenges}.
With the availability of large-scale pure-language data, such as posts on social media or forums, scripts of movies or TV series, and datasets from collection or crowdsourcing \cite{lison2016opensubtitles2016,henderson2019repository,chen2020jddc}, open-domain dialogue generation has developed rapidly.

\begin{figure}[!t]
\centering
\includegraphics[width=\linewidth]{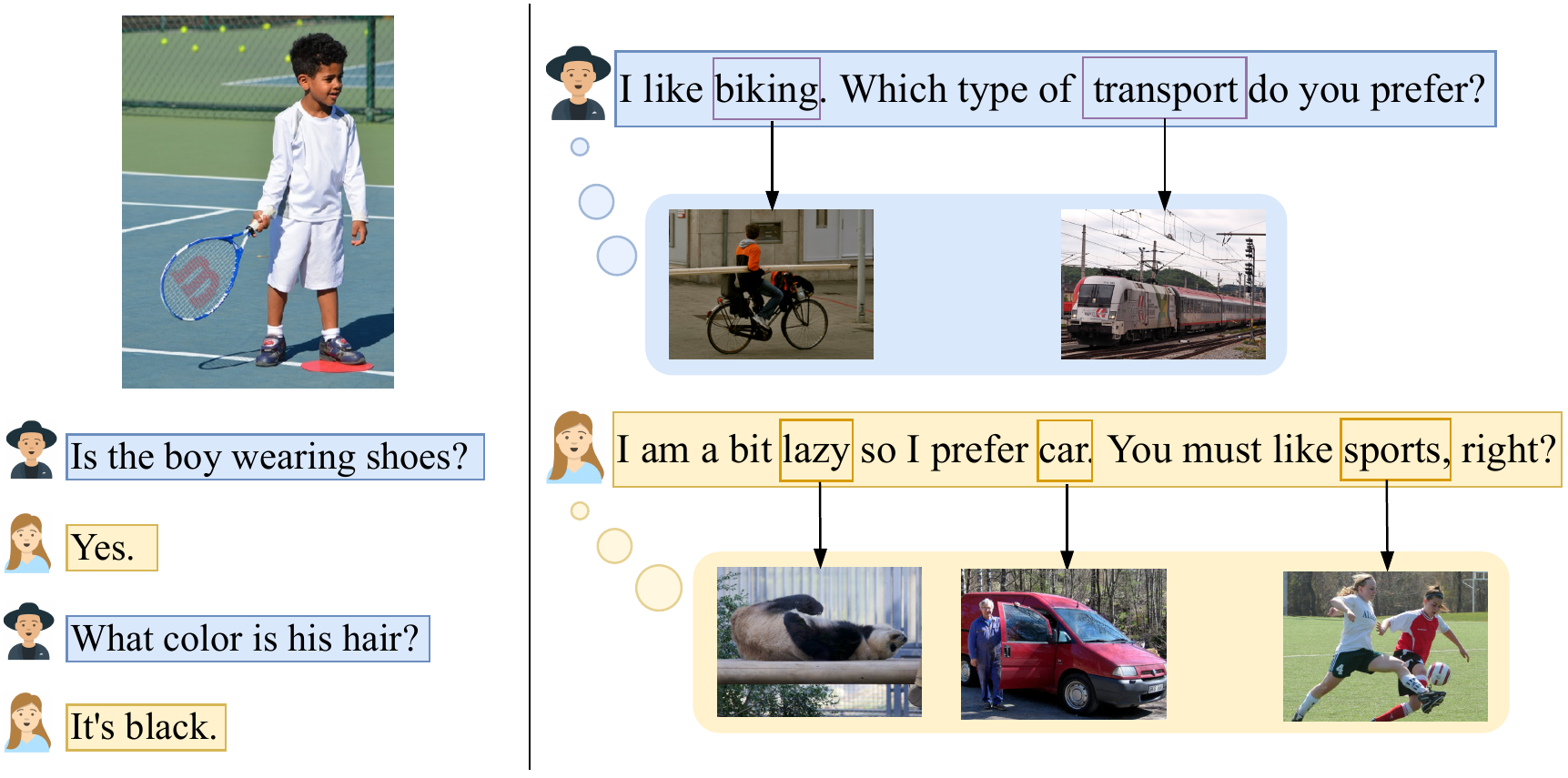}
\vspace{-5mm}
\caption{The comparison between visually-grounded Visual Dialogue (left) and pure-language Open-domain Dialogue (right). Left: Most visually-grounded dialogues talk about the content in a static image. Right: For pure-language dialogues, hidden images, named as visual impressions, should be explored and constructed, which enhances both dialogue understanding and generation.}
\label{fig:intro_case}
\end{figure}

However, only taking text-only data into consideration is not sufficient to produce a human-like system, since human communication is a multimodal interaction that consists of perception, construction, and exchange procedures. Humans have the ability to picture their understanding of the dialogue contents in their minds as a conversation goes on. These images, named as \textit{visual impressions}, mainly reflect some scenarios or objects that are tightly related to the conversation, and can be used to assist the response generation. Generally, visual impressions are hidden in text-only datasets, thus few work shed light on constructing or utilizing them for better open-domain dialogue understanding and generation. 

Previous studies have introduced visual information into dialogue systems, for example, Visual Dialogue (VisDial) and Multimodal Dialogue (MMD). The former is visually-grounded and needs to answer a series of questions based on a pre-given image and the dialogue history \cite{wang2020vd}, while the latter aims to fulfill a task (i.e., task-oriented) and is mainly constrained in a specific domain, e.g., fashion. However, open-domain dialogues are not grounded on certain of pre-defined images and usually contain contents in various domains. Figure \ref{fig:intro_case} shows two examples of VisDial and Open-domain Dialogue. Therefore, there is a large discrepancy between the visually-grounded/domain-specific datasets and the pure-language datasets for open-domain dialogue generation \cite{tan2020vokenization}. In addition, both VisDial and MMD require parallel image-dialogue data that is manually annotated, which is time-consuming and leads to small and simple datasets \cite{das2017visual,chauhan2019ordinal,saha2018towards}. 

Besides, almost each post-response pair\footnote{In this paper, we focus on the single-turn open-domain dialogue generation to illustrate our motivation and method concisely. Our approach will also work in the multi-turn setting when we replace the post with dialogue history, i.e., context.} in a open-domain dialogue has following two features: (1) They do not share the same semantic space, and topic transition often occurs; (2) Rather than word-level alignments, utterance-level semantic dependency exists in each pair. Therefore, when integrating visual impressions into open-domain dialogue generation, we need to take advantage of both post visual impressions (PVIs) and response visual impressions (RVIs). To generate a more appropriate response, PVIs should be utilized in the encoder for better understanding of posts, while RVIs are supposed to work as the assistance in the decoder and avoid producing deviated responses. Given a post in the test set, its PVIs can be constructed correspondingly, but the RVIs are unavailable, which causes inconsistency in training and test processes.

To alleviate these problems, we propose a two-stage framework that constructs visual impressions (VIs) at first and then, incorporates these VIs into open-domain dialogue generation with a proposed model, \textbf{Vis}ual-\textbf{A}ided \textbf{D}ialogue Transformer (VisAD). Specifically, at the first stage, we employ an image captioning dataset \textit{Flickr30K} \cite{young2014image} to train a word-image mapping model. Then, the content words (words that have substantive meanings and are crucial to evoke visual impressions in human minds) in a post will be firstly extracted and then sent into the mapping model to get corresponding PVIs. At the second stage, the co-attention encoder in the VisAD model is used to encode textual and visual information. Since the RVIs are not given in the test set, the cascade decoder with two sub-decoders predicts the content words in response first, and then uses the mapping model to obtain RVIs. Finally, the RVIs are integrated into the response generation process. We conduct extensive experiments on two open-domain dialogue datasets \textit{DailyDialog} \cite{li2017dailydialog} and \textit{PersonaChat} \cite{zhang2018personalizing}. Experimental results on both automatic and human evaluations show that our proposed framework outperforms competitive baselines over most of the metrics. 
The main contributions of this paper are listed as follows:
\begin{itemize}
    \item To the best of our knowledge, this paper is the first work to integrate visual information into pure-language open-domain dialogue generation. We propose a two-stage framework to break the constraint of parallel image-dialogue datasets and enable visual information to benefit text-only dialogue understanding and generation. 
    \item Since the post and response information are not in the same semantic space, we introduce post visual impressions (PVIs) and response visual impressions (RVIs), and further design a VisAD model to utilize PVIs and RVIs in the co-attention encoder and cascade decoder, respectively.
    \item Experimental results confirm the effectiveness of our proposed method with very favorable performance over the baselines in terms of fluency, relatedness and diversity.
\end{itemize}

\section{Related Work}

Our work is not only related to the traditional open-domain dialogue in pure language, but also some vision-language tasks, such as multimodal machine translation \cite{zhang2020neural}, multimodal summarization \cite{zhu2018msmo}, text to image generation \cite{xu2018attngan}, image captioning \cite{vinyals2015show}, visual question answering \cite{liang2018focal}, visual dialogue \cite{das2017visual,shuster2020image}, and multimodal dialogue \cite{saha2018towards}. 

\subsection{Open-domain Dialogue}
The so-called dialogue systems in natural language processing (NLP) can be roughly categorized into two groups: task-oriented and open-domain dialogue systems. Task-oriented dialogue systems are designed to complete tasks in specific domains, such as flight booking, hotel reservation, and customer service \cite{huang2020challenges,liu2020nlpcc}. Open-domain dialogue systems, however, aim to converse with humans in different domains and open-ended topics. To build an open-domain dialogue agent, existing methods are mainly retrieval-based or generation-based. The former tries to retrieve the appropriate responses from a pre-collected corpus, while the latter, also known as open-domain dialogue generation, utilizes the data-driven encoder-decoder framework to generate a response from the scratch.

Generating a coherent and informative response given a  pure-language post is a challenging task, since (1) more extra information is needed to support the post understanding and response generation, (2) each post and its response are usually not in the same semantic space, and (3) almost no word alignments exist in a post-response pair. Several methods \cite{zhao2017learning,serban2017hierarchical,shen2019modeling,ke2018generating,li2016persona,zhou2018emotional,meng2019refnet,zhou2018commonsense,shen2021icassp,zhan2021naacl} have been proposed to tackle these issues. For example, \citet{lian2019learning} used posterior distribution~\cite{zhan2020user,zhan2021probing} to select relevant knowledge for response generation. \citet{shen2020cdl} employed emotion labels to generate emotion-controllable responses. \citet{xing2017topic} imported a pre-trained topic model to predict the topic words in response. Other works \cite{luo2018auto,shen2021acl} utilize  auto-encoder and latent variables to learn utterance-level semantic dependency. 

To the best of our knowledge, our work takes the first attempt to integrate visual information, i.e., visual impressions, into the pure-language open-domain dialogue generation task.

\subsection{Visual Dialogue}
The visual dialogue task was proposed by \citet{dasvisual}, and requires an agent to answer multi-round questions about a static image \cite{das2017visual,ijcai2020DAM,huber2018emotional}. Previous work \cite{wu2018question,kottur2018visual,yang2019making,guo2019image,niu2019knowledge,kang2019dual,jiang2020dualvd} focused on developing different attention mechanisms to model the interactions among image, question, and dialogue history \cite{wang2020vd}. For example, \citet{gan2019multi} proposed a multi-step reasoning approach to answer a series of questions about an image with the recurrent dual attention mechanism. Recently, vision and language pre-training that aims to build joint cross-modal representations has attracted lots of attention from researchers \cite{li2019visualbert,lu2019vilbert,tan2019lxmert,su2019vl,alberti2019fusion,li2020unicoder}. Models based on Transformer encoder are designed for visually-grounded tasks and yield prominent improvement mainly on vision-language understanding. \citet{wang2020vd} proposed VD-BERT model that relies on the self-attention mechanism within a single-stream architecture to capture above mentioned interactions for visual dialogue.

However, since these visually-grounded models only utilize one image at a time, and few of them can handle the generative setting, they are not suitable for open-domain dialogue generation with multiple visual impressions integrated in a post or response. 

\subsection{Multimodal Dialogue}
\citet{saha2018towards} constructed a Multimodal Dialogue (MMD) dataset for the fashion domain, which consists of over 150K conversations with domain knowledge. Besides, they also presented a basic Multimodal Hierarchical Encoder-Decoder model (MHRED). Based on MHRED, following studies focused on either understanding and fusing multimodal semantics or incorporating the domain knowledge properly \cite{liao2018knowledge,chauhan2019ordinal,cui2019user,nie2019multimodal}. For example, \citet{cui2019user} devised a User attention-guided Multimodal Dialogue (UMD) model that paid more attention to the user requirements and encoded the dialogue history dynamically based on users’ attention. \citet{liao2018knowledge} stored style tips knowledge into memory networks and employed an attention mechanism to select useful knowledge. Recently, \citet{he2020multimodal} proposed a novel transformer-based model that can deal effectively with the dependencies between multimodal semantic elements, and reached the best result on the MMD dataset.

The existing multimodal dialogue systems are solely based on the MMD dataset, where all conversations are constrained in the fashion domain and mainly aim to fulfill specific tasks, such as product recommendation. Our work focuses on open-domain dialogue generation with various domains, and the constructed visual impressions reflect scenarios or objects related to the dialogue contents based on human understanding, rather than fashion images for shoes or dresses.

\subsection{Vision Utilization for Pure-language Tasks}

Recently, some studies focused on building a unified visually-aided framework to improve the pure-language tasks with external visual information. \citet{tan2020vokenization} added images to improve general language understanding and used image indices as labels in the discriminative setting. \citet{zhang2020neural} utilized images for pure-language neural machine translation in the generative setting. Due to the shared semantic space of source and target sentences, the image indicted by each source-target sentence pair is the same.

In contrast to the above mentioned work, our visually-aided model is designed in the generative setting and employs both PVIs and RVIs, which is more appropriate for open-domain dialogue generation as described in Section \ref{sec:intro}.

\section{Method}

\begin{figure}[!t]
\centering
\includegraphics[width=0.8\linewidth]{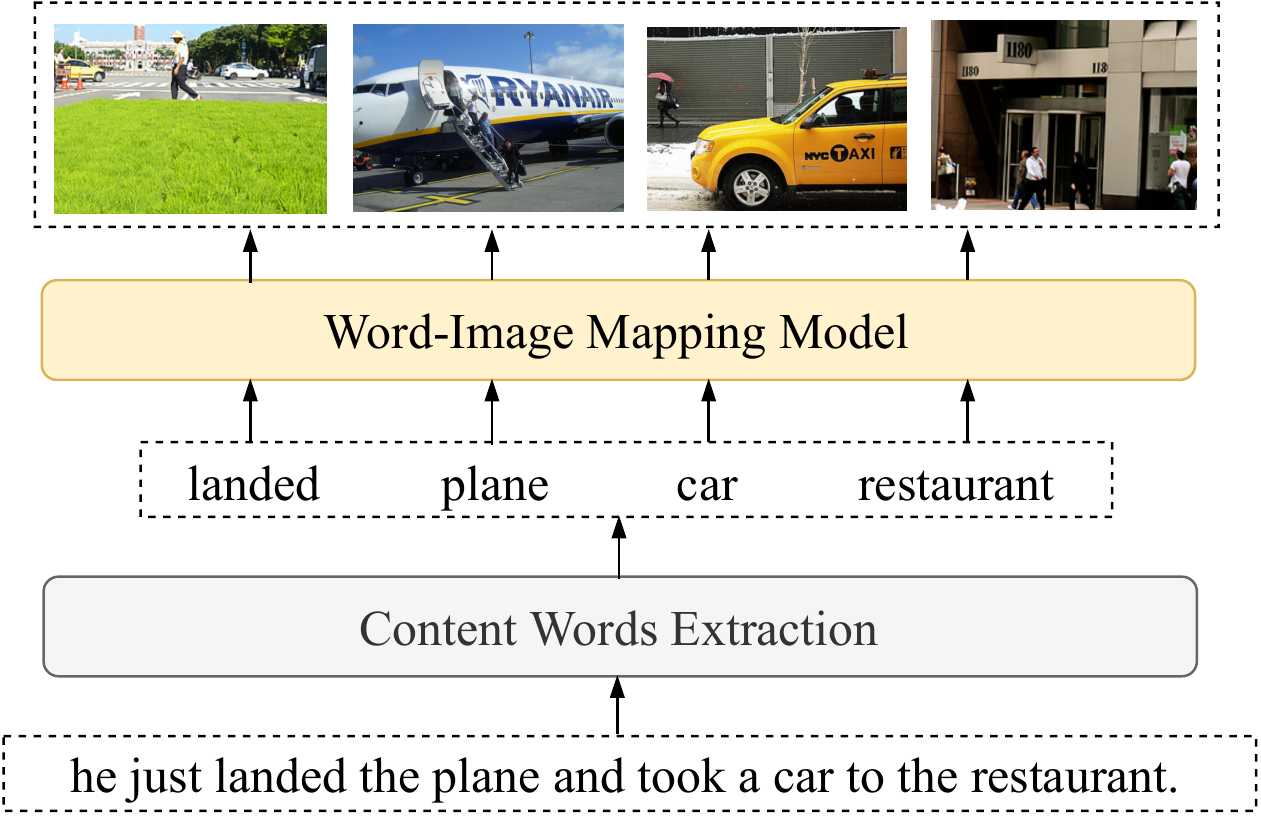}
\vspace{-3mm}
\caption{The process of constructing visual impressions.}
\label{fig:exper-lookup}
\end{figure}

In this section, we firstly introduce our task formulation for open-domain dialogue generation with visual impressions (Section \ref{task-def}), and then illustrate the proposed two-stage framework. At the first stage (Section \ref{kw-map}), we employ an image captioning dataset to train a word-image mapping model, which will further be used to match an utterance\footnote{In this paper, ``sentence'' and ``utterance'' have the same meaning, and are used separately in the image captioning and dialogue generation task.} with corresponding images. At the second stage (Section \ref{visad}), we propose a model VisAD with the co-attention encoder and cascade decoder to incorporate PVIs and RVIs for both post understanding and response generation, respectively.

\subsection{Task Formulation}
\label{task-def}
Given the post $X=\{x_i\}^{|X|}_{i=1}$ as input, where $x_i$ is the $i$-th token and $|X|$ is the length of $X$, the pure-language open-domain dialogue model needs to generate a response as: $\hat{R} = \mathop{\arg\max} \mathcal{P}(R|X)$.

We further utilize visual impressions, and reformulate the generation process as following steps: (1) Given a post, obtaining its PVIs with the word-image mapping model which will be explained in a minute. (2) Predicting content words that will be used in response based on the post and its PVIs, and obtain the RVIs with the generated content words. (3) Generating the response based on the post, PVIs, and RVIs: $\hat{R} = \mathop{\arg\max} \mathcal{P}(R|X, PVI, RVI)$.

\subsection{First Stage: Visual Impression Construction}
\label{kw-map}
Following procedures are implemented on an image captioning dataset \textit{Flickr30K} \cite{young2014image}), in which each image $v$ corresponds to a sentence $s$ that describes the visual content in $v$. Since both image captioning and open-domain dialogue generation are in open-ended domains, we apply a mapping model to capture the semantic relationship in the parallel image-sentence dataset and utilize the relations to import visual impressions for utterances in pure-language datasets. The process of this stage is shown in Figure \ref{fig:exper-lookup}.

\paragraph{\textbf{Content Word Extraction}} 
In linguistics, words in a sentence can be categorized into two types: content words (words that have substantive meanings and are crucial to evoke visual impressions in human minds) and function words (words that play a functional role in making a natural and grammatical sentence) \cite{hill1952structure}. For each image-sentence pair in the image captioning dataset, we apply a rule-based
content word extractor to automatically extract content words from the sentence. We remove stopwords, and then obtain the Part-Of-Speech (POS) feature for each word. As a content word, its POS tagging should be noun, verb, adjective or adverb \cite{kong2020tsdg}. 

\begin{figure*}[t]
\centering
\includegraphics[width=0.95\linewidth]{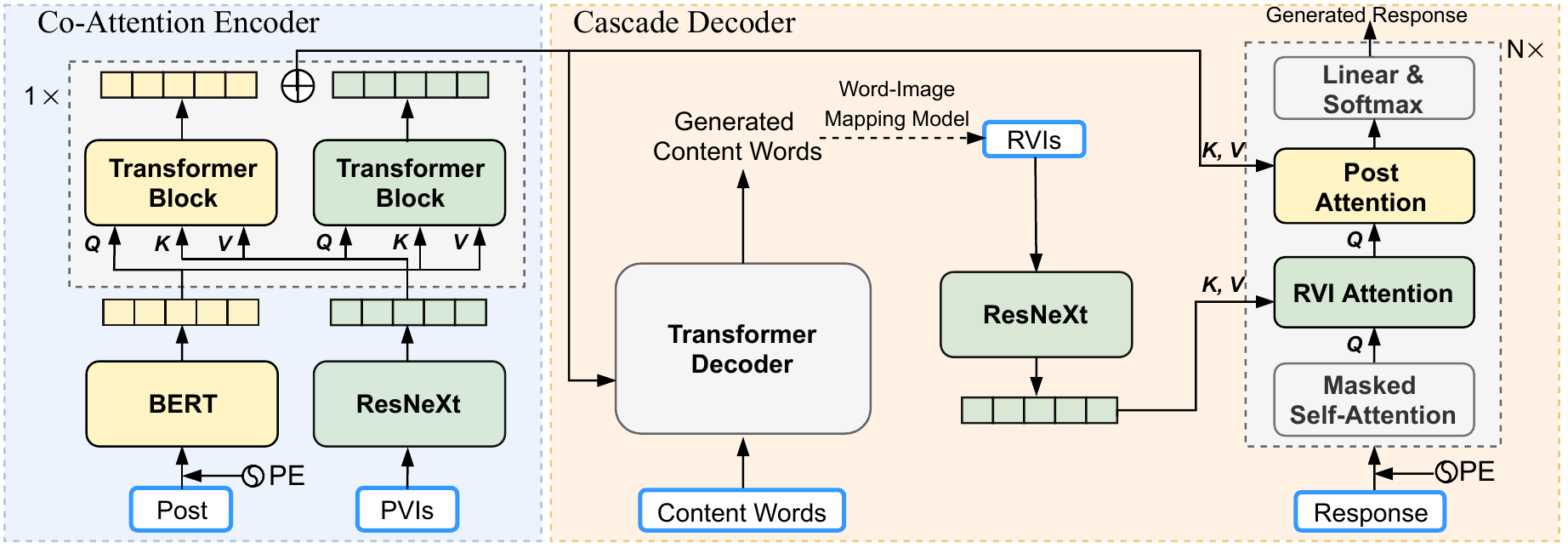}
\caption{The architecture of VisAD. For brevity, we omit the residual connection and layer normalization layers. We keep Transformer Block, Transformer Decoder, and other parts the same as the original Transformer \cite{vaswani2017attention}.}
\label{fig:visad}
\end{figure*}

\paragraph{\textbf{Word-Image Mapping Model}} 
The purpose of applying this model is that given an utterance (the post or response), its visual impressions can be returned and further utilized by the second stage. In an image captioning dataset, the linguistic words of a sentence $s$ are mainly aligned with the semantic units (object, object attribute, and object relationship)\cite{guo2019aligning} in the grounded image $v$. Therefore, the image $v$ matches several content words $\{c_i\}^k_{i=1}$ in the sentence $s$. The word-image mapping model takes a content word $c_i$ and an image $v$ as input, and the output $\mathrm{Sim}(c_i, v)$ is the relevance score between content word $c_i \in s$ and image $v$. We apply the widely used cosine similarity to measure the relevance between the content word representation $\textbf{H}_{i}^{c}$ and image representation $\textbf{H}^{v}$:
\begin{equation}
   \mathrm{Sim}(c_i, v) = cos(c_i, v) = \frac{{\textbf{H}_{i}^{c}} \cdot \textbf{H}^v}{|\textbf{H}_{i}^{c}| |\textbf{H}^v|}.
\end{equation}

Here, we employ two pre-trained models to get word and visual representations, respectively. For word representation, we feed the sentence (represented as a sequence of content words $s=\{c_i\}^k_{i=1}$) into a pre-trained BERT model \cite{devlin2019bert}, and get the output vector ${\mathbf{h}_i^c}$ for word $c_i$. At the same time, the initial visual representation $\mathbf{h}^{v}$ for a image $v$ is embedded by a pre-trained ResNeXt \cite{xie2017aggregated}. Then, we apply two multi-layer perceptions (MLPs) to map the word and image embeddings into the same space, and obtain their final representations:
\begin{equation}
    \textbf{H}_{i}^{c} = {\rm MLP}^{c}(\mathbf{h}_i^c), \quad \textbf{H}_{v} = {\rm MLP}^{v}(\mathbf{h}^v).
\end{equation}

\paragraph{\textbf{Training}} 
In order to train the word-image mapping model, we need to construct some negative word-image instances. We first get the negative sentence-image pairs by randomly sampling another different image $v'$ from the same dataset for the sentence $s$. Then, $v'$ and each content word $c_i$ in $s$ compose a negative word-image instance. Finally, the training objective of this word-image mapping model can be formulated as:
\begin{equation}
  \mathop{\arg\min}\sum_{i=1}^{k}\mathrm{max}\{0, - \mathrm{Sim}(c_i, v) + \mathrm{Sim}(c_i, v') + thr \},
\end{equation}
where $thr$ is a threshold margin, which indicates that the relevant score of positive instances $\mathrm{Sim}(c_i, v)$ should be larger than the negative ones $\mathrm{Sim}(c_i, v')$ by at least a threshold $thr$. Therefore, minimizing the loss function is able to force the word-image mapping model to discriminate positive and negative instances.

\paragraph{\textbf{Inference}}
Given the open-domain dialogue datasets, to map the content words in each utterance with related visual impressions, we need to utilize the well-trained word-image mapping model\footnote{Though the model is trained on the image captioning dataset, the images and sentences are also in open-ended domains. Therefore, it is suitable to apply it in the open-domain dialogue datasets, and we leave the problem of transfer learning and domain adaptation for future work.}. As shown in Figure \ref{fig:exper-lookup}, given an utterance, content words in it will be firstly extracted through the content word extraction process. Then, these words will be sent into word-image mapping model individually to retrieve their related images. We return the retrieved top-1 image from a set of images $V$ regarding the word-image relevance score\footnote{We store the obtained word-image table in advance to accelerate the training and test processes of the cascade decoder.}. Finally, an utterance with $k$ content words will be matched with $k$ images, denoted as $\mathrm{VI} = \{v_1, v_2, ..., v_k\}$.

\subsection{Second Stage: Response Generation}
\label{visad}

In this sub-section, we will describe the architecture of our proposed model, VisAD, as shown in Figure \ref{fig:visad}. To better understand the dialogue post, make use of the relations between post/PVIs and response/RVIs, and further assist the response generation process, VisAD consists of a co-attention encoder and a cascade decoder that is composed of two sub-decoders: content word decoder and response decoder.

\paragraph{\textbf{Co-Attention Encoder}}

Given a post $X=\{x_i\}^{|X|}_{i=1}$, the word embedding and position embedding vectors are represented as $\mathrm{WE}(X)$ and $\mathrm{PE}(X)$, respectively. The initial post representation $\mathbf{E}_{X}$ is computed through a pre-trained BERT \cite{BERT2019} model: 
\begin{align}
    &\mathbf{e}_{X} = \mathrm{WE}(X) + \mathrm{PE}(X),
    &\mathbf{E}_{X} = {\rm BERT}_{base}(\mathbf{e}_{X}).
\end{align}

At the same time, we employ ResNeXt \cite{xie2017aggregated} to transform PVIs into an image representation $\mathbf{E}_{PVI}$, which is defined as follows:
\begin{align}
    &\mathbf{e}_{PVI}^{i} = {\rm ResNeXt}(v_i) + {\rm PE}(v_i), \\
    &\mathbf{E}_{PVI} = {\rm Concat}(\mathbf{e}_{PVI}^{1},\mathbf{e}_{PVI}^{2},...,\mathbf{e}_{PVI}^{k}) \label{keq},
\end{align}
where $k$ is the number of PVIs related to the post $X$, $v_i$ is the $i$-th PVI, ${\rm PE}(v_i)$ is its position embedding regarding the entire sequence of PVIs, and $\mathbf{e}_{PVI}^{i}$ is its representation. 

Then, we utilize Transformer Block \cite{vaswani2017attention} to implement the co-attention between post representation $\mathbf{E}_{X}$ and its PVI representation $\mathbf{E}_{PVI}$. A Transformer Block consists of a multi-head attention sub-layer (MHA$(\cdot)$) and a position-wise feed-forward network (FFN$(\cdot)$). We denote a Transformer Block as $\mathrm{TB}$(\textbf{Q}, \textbf{K}, \textbf{V}), where $\textbf{Q}$, $\textbf{K}$, and $\textbf{V}$ correspond to the inputs of an MHA layer. Please refer to \citet{vaswani2017attention} for more details. Then, the image-enhanced post representation $\widetilde{\mathbf{E}}_{X}$ and text-enhanced PVI representation $\widetilde{\mathbf{E}}_{PVI}$ are defined as:
\begin{align}
    &\widetilde{\mathbf{E}}_{X} = \mathrm{TB}(\mathbf{E}_{X}, \mathbf{E}_{PVI}, \mathbf{E}_{PVI}), \\
    &\widetilde{\mathbf{E}}_{PVI} = \mathrm{TB}(\mathbf{E}_{PVI}, \mathbf{E}_{X}, \mathbf{E}_{X}).
\end{align}

Finally, we concatenate the image-enhanced post representation and text-enhanced PVI representation to get an unified representation $\widetilde{\mathbf{U}}$:
\begin{equation}
    \widetilde{\mathbf{U}} = {\rm Concat}(\widetilde{\mathbf{E}}_{X}, \widetilde{\mathbf{E}}_{PVI}).
\end{equation}

\paragraph{\textbf{Content Word Decoder}}
The content word decoder aims to predict the content words in response by considering the post and its PVIs. More importantly, the generated content words can be used to help retrieve RVIs that will be incorporated in the response decoder for response generation. 

During training, the ground-truth content words in a response can be firstly extracted by the content word extraction part (Section \ref{kw-map}). We then formulate the content word prediction as a generation task\footnote{Content word generation and prediction have the same meaning in this paper.} and utilize the original Transformer decoder \cite{vaswani2017attention} to implement it. Given the concatenated vector $\widetilde{\mathbf{U}}$ from the encoder and generated content words before $i$-th step, the probability of generating the $i$-th content word $c_i$ can be calculated as:
\begin{equation}
   p(c_i|X, PVI, c_{<i};\theta) = {\rm TD}(\widetilde{\mathbf{U}}, \mathbf{E}_{c<i}),
\end{equation}
where $\mathrm{TD}(\cdot)$ is the original Transformer decoder \cite{vaswani2017attention}, and $\mathbf{E}_{c<i}$ denotes the embeddings of content words before $c_i$.

Therefore, the training objective of this decoding stage is defined as: 
\begin{equation}
    \mathcal{L}_{c}(\theta) = - \sum_{i=1}^{T_{c}}{\mathrm{log}}(p(c_i|X, PVI, c_{<i};\theta)).
\end{equation}

\paragraph{\textbf{Response Decoder}}
In the second decoding phase, we propose a RVI-aware decoder to utilize the post and RVIs for response generation. As shown in Figure \ref{fig:visad}, RVIs for the response can be obtained according to the predicted content words in the first decoding stage (for training process, we use the ground-truth content words to get related RVIs). The initial RVI representation $\mathbf{E}_{RVI}$ is computed as:
\begin{align}
    &\mathbf{e}_{RVI}^{i} = {\rm ResNeXt}(v_i) + {\rm PE}(v_i), \\
    &\mathbf{E}_{RVI} = {\rm Concat}(\mathbf{e}_{RVI}^{1},\mathbf{e}_{RVI}^{2},...,\mathbf{e}_{RVI}^{n}) \label{neq},
\end{align}
where $v_i$ is the $i$-th image of the RVI sequence and it is related to the $i$-th content word, $n$ is the number of RVIs.

At the $t$-th decoding step, previous generated words $R = r_{<t}$ are represented using a masked self-attention layer:
\begin{align}
    &\mathbf{e}_{R} = \mathrm{WE}(R) + \mathrm{PE}(R),
    &\mathbf{E}_{R} = \mathrm{MHA}(\mathbf{e}_{R}, \mathbf{e}_{R}, \mathbf{e}_{R}).
\end{align}
After getting the initial representations, obtained RVIs will be incorporated through RVI attention layer (the green block in the response decoder in Figure \ref{fig:visad}). Then, the RVI-aware representation is defined as:
\begin{equation}
    \widetilde{\mathbf{E}}_{R} = \mathrm{MHA}(\mathbf{E}_{R}, \mathbf{E}_{RVI}, \mathbf{E}_{RVI}).
\end{equation}
The response representation $\widetilde{\mathbf{E}}_{R}^{'}$ that incorporates the unified representation $\widetilde{\mathbf{U}}$ from the encoder is formulated as:
\begin{equation}
   \widetilde{\mathbf{E}}_{R}^{'} = \mathrm{MHA}(\widetilde{\mathbf{E}}_{R}, \widetilde{\mathbf{U}}, \widetilde{\mathbf{U}}).
\end{equation}

Finally, the decoding distribution over the vocabulary for the $t$-th word is computed as:
\begin{equation}
    p(r_t|X, PVI, RVI, r_{<t}) = {\rm Softmax}({\rm Linear}(\widetilde{\mathbf{E}}_{R}^{'})),
\end{equation}
where $\mathrm{Linear}(\cdot)$ and $\mathrm{Softmax}(\cdot)$ are the linear transformation layer and Softmax activation function, respectively, and we use the same definition as the original Transformer model \cite{vaswani2017attention}.

\paragraph{\textbf{Training objective}}
Our training objective is composed of two parts: the loss from the first and second decoding phases. Similar to previous work, the training loss of the response generation process is computed as:
\begin{equation}
    \mathcal{L}_{nll}(\theta) = -\sum_{t=1}^{T}{\mathrm{log}}(p(r_t|, X, PVI, RVI, r_{<t};\theta)).
\end{equation}

Finally, the total loss function is defined as:
\begin{equation}
    \mathcal{L}_{full}(\theta) = \mathcal{L}_{nll}(\theta) + \alpha\mathcal{L}_c(\theta),
\end{equation}
where $\alpha$ is a hyper-parameter to control the trade-off between the content word generation and response generation. $\theta$ refers to parameters of the entire model and is optimized in an end-to-end framework.
\section{Experiments}

\begin{table*}[!t]
\caption{The results of automatic evaluation on \textit{DailyDialog} and \textit{PersonaChat} datasets. The metrics Perplexity, BLEU, Distinct-1/2, Average, Extrema, and Greedy  are abbreviated as PPL, B, D1, D2, Avg, Ext, and Gre, respectively. The best results are highlighted with \textbf{bold}. ``$^\star$'' denotes that the result is statistically significant with $p < 0.01$.}
\vspace{-3mm}
\centering
\resizebox{\textwidth}{17mm}{
\begin{tabular}{l|ccccccc|ccccccc}
\toprule[1pt]
\textbf{Dataset} & \multicolumn{7}{c|}{\textbf{DailyDialog}} & \multicolumn{7}{c}{\textbf{PersonaChat}} \\ \cline{1-15}
 \textbf{Model} & \textbf{PPL} & \textbf{B} & \textbf{D1} & \textbf{D2} & \textbf{Avg} & \textbf{Ext} & \textbf{Gre} & \textbf{PPL} & \textbf{B} & \textbf{D1} & \textbf{D2} & \textbf{Avg} & \textbf{Ext} & \textbf{Gre} \\
   \hline
   SEQ2SEQ & 42.53 & 2.85  & 4.83 & 17.27 & 50.21 & 35.28 & 43.77  & 39.62 & 0.97  & 2.86 & 11.26 & 54.36 & 35.58 & 46.30\\
   CVAE & 44.23 & 5.32  & 6.52 & 19.56 & 52.60 & 33.55 & 42.82  & 42.22 & 1.68  & 3.94 & 15.48 & 55.02 & 34.71 & 47.91\\
   DialogWAE  & 35.47 & 6.07  & 5.95 & 19.28 & 51.39 & 38.17 & 45.16  & 33.26 & 1.57  & 4.08 & 17.62 & 42.38 & 35.74 & 48.89 \\
   Transformer & 29.36 & {6.93}  & 5.62 & 21.85 & 48.69 & 36.67 & 45.59  & 27.81 & 4.61 & 5.46 & 23.19 & 53.69 & 38.69 & 50.69   \\
   GVT & 26.69 & 5.86  &  5.59 & 20.06 & 57.82 & 41.30 & 48.37 & 28.89 & 3.82  & 5.59 & 25.16 & 58.47 & 39.15 & 46.24  \\
   VD-BERT & 27.05 & 4.93  &  6.31 & 24.17 & 55.61 & 39.92 & 46.15 & 25.11 & 4.97  & 4.86 & 20.71 & 56.23 & 36.08 & 47.04  \\ \cline{1-15}
   \textbf{VisAD} & \textbf{17.81$^\star$} & \textbf{12.47$^\star$}  & \textbf{9.68$^\star$} & \textbf{33.22$^\star$} &\textbf{64.21$^\star$} & \textbf{45.19$^\star$} & \textbf{52.56$^\star$}  & \textbf{19.37$^\star$} & \textbf{9.08$^\star$}  & \textbf{12.70$^\star$} & \textbf{34.78$^\star$} & \textbf{62.27} & \textbf{46.02$^\star$} & \textbf{53.88$^\star$}   \\
\bottomrule[1pt]
\end{tabular}}
\label{tab_auto-eval}
\end{table*}

\subsection{Datasets}
We evaluate our model on two open-domain dialogue datasets, \textit{DailyDialog} \cite{li2017dailydialog} and \textit{PersonaChat} \cite{zhang2018personalizing}. \textit{DailyDialog} has 13,118 daily conversations, while \textit{PersonaChat} contains 10,981 dialogues and has a set of 1,155 personas. We use the original version of \textit{PersonaChat} and ignore those personas. The datasets are separated into training, validation, and test sets with the same ratios as in the baseline papers. While the raw datasets are arranged in multi-turn sessions, we consider each of two consecutive utterances as a post-response pair. In our settings, the number of pairs in training/validation/test set is 33,776/2,690/1,000 for \textit{DailyDialog}, and 39,934/3,928/1,000 for \textit{PersonaChat}.
In order to train the word-image mapping model, we also utilize an image captioning dataset \textit{Flickr30K} \cite{young2014image} that consists of 31,783 sentence-image pairs. These images are also used as candidates ($V$ in Inference part of Section \ref{kw-map}) to obtain related visual impressions for dialogue utterances.

\begin{table}[t]
\caption{The results of human evaluation on (a) \textit{DailyDialog} and (b) \textit{PersonaChat} datasets.}
\vspace{-3mm}
\centering
\begin{tabular}{c|l|ccc|c}
  \toprule[1pt] 
  ~ & \textbf{Opponent} & \textbf{Win} & \textbf{Loss} & \textbf{Tie} & \textbf{Kappa}  \\
  \hline
  \multirow{6}{*}{(a)} & VisAD vs. SEQ2SEQ & 65\% & 18\% & 17\% & 0.628 \\
   & VisAD vs. CVAE & 53\% & 21\% & 26\% & 0.514 \\
   & VisAD vs. DialogWAE & 44\% & 30\% & 26\% & 0.463  \\
   & VisAD vs. Transformer & 54\% & 29\% & 17\% & 0.560   \\
   & VisAD vs. GVT & 41\% & 30\% & 29\% & 0.439  \\
   & VisAD vs. VD-BERT & 55\% & 23\% & 22\% & 0.576    \\ \hline
  \multirow{6}{*}{(b)} & VisAD vs. SEQ2SEQ &  57\% & 20\% & 23\% & 0.604\\
   & VisAD vs. CVAE & 45\% & 27\% & 28\% & 0.468 \\
   & VisAD vs. DialogWAE & 53\% & 21\% & 26\% & 0.540 \\
   & VisAD vs. Transformer &  40\% & 31\% & 29\% & 0.427 \\
   & VisAD vs. GVT & 48\% & 27\% & 25\% & 0.506 \\
   & VisAD vs. VD-BERT & 55\% & 19\% & 26\% & 0.559  \\
  \bottomrule[1pt]
\end{tabular}
\label{tab:human}
\end{table}

\subsection{Implementation Details} 
At the first stage, to train the word-image mapping model, we employ the pre-trained ResNeXt \cite{xie2017aggregated} model and $\mathrm{BERT}_{base}$ \cite{devlin2019bert} model with 768 dimensions to get the initial representations of images and content words, respectively. The size of image set $V$ is 31,783. At the inference stage for word-image mapping model, content words of each utterance in dialogue datasets will be firstly extracted and then mapped to related visual impressions. For each content word $c_i$, we keep the image with the highest relevance score $\mathrm{Sim}(c_i, v)$ as the visual impression $v_i$. 
Regarding our model VisAD, the maximum decoding length of the response is set to 50. The number of content words/visual impressions for each dialogue utterance is set to 8 (i.e., both $k$ in Equation \ref{keq} and $n$ in Equation \ref{neq} equal to 8). The vocabulary size is set to 20,000. The dimension of word embeddings is set to 512. The dropout rate and smoothing factor in $\mathrm{Softmax}$ are set to 0.1. The Adam optimizer with a learning rate of 0.001 is used to train the model with batch size of 128. The $\beta_{1}= 0.9$ and $\beta_{2}= 0.998$ are used for gradient optimization. $\alpha$ in the total loss function equals to 0.5. The entire model is trained on four Tesla P40 GPUs. 

\subsection{Baseline Models} 
We compare our model with following representative baselines:
    (1) \textbf{SEQ2SEQ} \cite{bahdanau2015neural}: the vanilla sequence-to-sequence model with cross-attention mechanism. 
    (2) \textbf{CVAE} \cite{zhao2017learning}: a conditional variational auto-encoder model.
    (3) \textbf{DialogWAE} \cite{gu2018dialogwae}: a wasserstein auto-encoder model. The input context in the original model is the post in our paper. 
    (4) \textbf{Transformer} \cite{vaswani2017attention}: a sequence-to-sequence model with self-attention mechanism. 
    (5) \textbf{GVT} \cite{lin2020variational}: a transformer-based model with a global latent variable. 
    (6) \textbf{VD-BERT} \cite{wang2020vd}: a pre-trained model for visually-grounded dialogue generation. Since only one image can be utilized in the model, we just feed the first image of the PVI sequence, i.e., visual impression of the first content word in the post, into the model.
The first five models are designed for pure-language dialogue generation. Besides, GVT and VD-BERT are the state-of-the-art models for pure-language and visually-grounded dialogue generation, respectively.

\subsection{Evaluation Measures}
\paragraph{\textbf{Automatic Metrics}}
We adopt four groups of automatic metrics to assess the performance of dialogue generation models\footnote{We employ a popular NLG evaluation project available at \url{https://github.com/Maluuba/nlg-eval} for automatic evaluation. The final BLEU score is reported as the average value of BLEU-1 to BLEU-4 calculated in this project.}:
    (1) \textbf{Perplexity} \cite{serban2015hierarchical}: Perplexity measures the high-level general quality of the generation model, and usually a relatively lower Perplexity value represents a more fluent responses.
    (2) \textbf{BLEU} \cite{papineni2002bleu}: BLEU measures how much a generated response contains n-gram overlaps with the ground-truth response.
    (3) \textbf{Embedding-based metrics} \cite{serban2017hierarchical,liu2016not}: Rather than calculating the token-level or n-gram similarity as the Perplexity and BLEU, we also use three metrics to compute the word embedding similarity between the generated response and the ground truth \cite{liu2016not}: (a) Greedy: greedily matching words in two utterances based on the cosine similarities between their embeddings, and averaging the obtained scores; (b) Average: cosine similarity between the averaged word embeddings in the two utterances; (c) Extrema: cosine similarity between the largest extreme values among the word embeddings in the two utterances.
    (4) \textbf{Diversity metrics} \cite{li2016diversity}: We apply {\it Distinct} to report the degree of diversity. Dist-1/2 is defined as the ratio of unique uni/bi-grams over all uni/bi-grams in generated responses.

\begin{table*}[t]
\caption{Ablation results on \textit{DailyDialog} and \textit{PersonaChat} datasets. }
\vspace{-3mm}
\centering
\begin{tabular}{l|ccccccc|ccccccc}
\toprule[1pt]
 \textbf{Dataset} & \multicolumn{7}{c|}{\textbf{DailyDialog}} & \multicolumn{7}{c}{\textbf{PersonaChat}} \\ \cline{1-15}
 \textbf{Model} & \textbf{PPL} & \textbf{B} & \textbf{D1} & \textbf{D2} & \textbf{Avg} & \textbf{Ext} & \textbf{Gre} & \textbf{PPL} & \textbf{B} & \textbf{D1} & \textbf{D2} & \textbf{Avg} & \textbf{Ext} & \textbf{Gre} \\
   \hline
   (1) 1DE-orig & 30.74 & 7.29 & 6.13 & 23.29 & 53.10 & 37.79 & 47.52 & 29.19 &5.61 & 6.72 & 21.44 & 54.63 & 40.55 & 50.81 \\
   (2) 1DE-PVI & 33.19 & 8.60 & 5.69 & 22.74 & 51.26 & 37.21 & 46.95 & 31.19 & 5.93 & 8.28 & 26.50 & 56.06 & \textbf{45.82} & 51.76 \\
   (3) 2DE-PVI & 26.57 & 10.92 & 7.03 & 27.19 & 55.02 & 40.32 & 51.75 & 23.99 & 7.46 & 10.48 & 29.33 & 62.04 & 41.71 & 50.99 \\
   (4) 2DE-CW & 19.27 & \textbf{13.48} &8.27 &29.50 & 59.27 & 43.18 & \textbf{54.10} & \textbf{19.04} & 8.61 & 11.60 & 27.29 & 59.73 & 43.27 & 52.39\\\hline
   \textbf{VisAD} & \textbf{17.81} & {12.47}  & \textbf{9.68} & \textbf{33.22} &\textbf{64.21} & \textbf{45.19} & {52.56}  & {19.37} & \textbf{9.08}  & \textbf{12.70} & \textbf{34.78} & \textbf{62.27} & {45.02} & \textbf{53.88} \\
\bottomrule[1pt]
\end{tabular}
\label{tab:ablation}
\end{table*}

\paragraph{\textbf{Human evaluation}}
For human evaluation, we first randomly sample 300 posts from the test set and get the corresponding responses generated from each model. Next, we send pairs of the post and generated response to six professional annotators without order. Annotators are then required to evaluate among ``Win'' (response$_1$ is better), ``Loss'' (response$_2$ is better) and ``Tie'' (they are equally good or bad) independently, considering three aspects: Fluency, Coherence, and Informativeness. Fluency assesses the grammatical correctness and readability of the generated responses; Coherence evaluates whether the generated responses can respond to their posts naturally; Informativeness indicates whether the generated responses are informative and not generic. Each comparison is conducted between two responses generated by VisAD and a baseline model, respectively.

\subsection{Experimental Results}
\paragraph{\textbf{Automatic Evaluation Results}}
As shown in Table \ref{tab_auto-eval}, among all baselines, CVAE and DialogWAE surpass the vanilla SEQ2SEQ due to specific the latent variables and the hierarchical structure. Transformer and GVT also improve the performance a lot, owing to their self-attention mechanism. Engaging with the visual information, VD-BERT performs better than other models for pure-language dialogue generation. VisAD exhibits competitive improvement of PPL and embedding-based metrics over baselines, implying that visual impressions can promote the response fluency and relevance. In terms of diversity, CVAE, DialogWAE, and GVT also achieve better performance comparing to SEQ2SEQ, whereas our model presents much larger improvements in Dist-1/2, e.g., improved by 9.05 on Dist-2 in \textit{DailyDialog}. The result shows that, under the assistance of visual impressions, our model is more adept at generating diverse responses. VisAD significantly outperforms the baselines on the majority of metrics with $p$-value $<$ 0.01.

\paragraph{\textbf{Human Evaluation Results}}
We also carry out the human evaluation through comparisons between our model VisAD and other baselines. For each case, given a post-response pair, two generated responses are provided, one is from our model and the other is from the compared model. Table \ref{tab:human} summarizes the results of human judgement on two datasets. The kappa scores indicate that the annotators come to a moderate agreement in the judgment. Not surprisingly, VisAD consistently outperforms all the compared models. Besides, we notice that VisAD exhibits significant improvements comparing with GVT and VD-BERT, the SOTA models described in the baseline part. We analyze the bad cases and find that VD-BERT still suffers from generating unrelated or generic responses, because it can only utilize one image in the post. Augmented with the visual impressions, both PVIs and RVIs, VisAD introduces a competitive boost in response quality, which is in line with the automatic evaluation, confirming the superior performance of our proposed model.

\subsection{Ablation Study} 
To examine the effectiveness of integrating visual impressions into pure-language response generation, we conduct model ablations by removing and replacing particular module from VisAD. Here, we have four variants: (1) 1DE-orig: replacing the entire cascade decoder with one original Transformer decoder. (2) 1DE-PVI: using only one decoder to directly generate the response, and its structure is the same as ``response decoder''. However, the RVI attention needs to be changed into PVI attention, since there is only one decoding phase, content words in response are not generated, and no RVIs can be obtained. (3) 2DE-PVI: still using the cascade decoder with two sub-decoders, but replacing RVI attention in the response decoder with PVI attention. (4) 2DE-CW: using the cascade decoder and replacing RVI attention in the response decoder with content word attention, i.e., paying attention to the generated content words in response rather than RVIs. 
Model (1), (2) and (3) can only utilize the post and PVIs, the comparisons between VisAD and them show that the post and response information are not in the same semantic space, and employing RVIs are more suitable for response generation. Model (4) uses content words in response rather than RVIs, while VisAD integrates corresponding RVIs, the results of them indicate that images can provide external information to increase the diversity of generated response.

\begin{figure*}[t]
\centering
\includegraphics[width=0.99\linewidth]{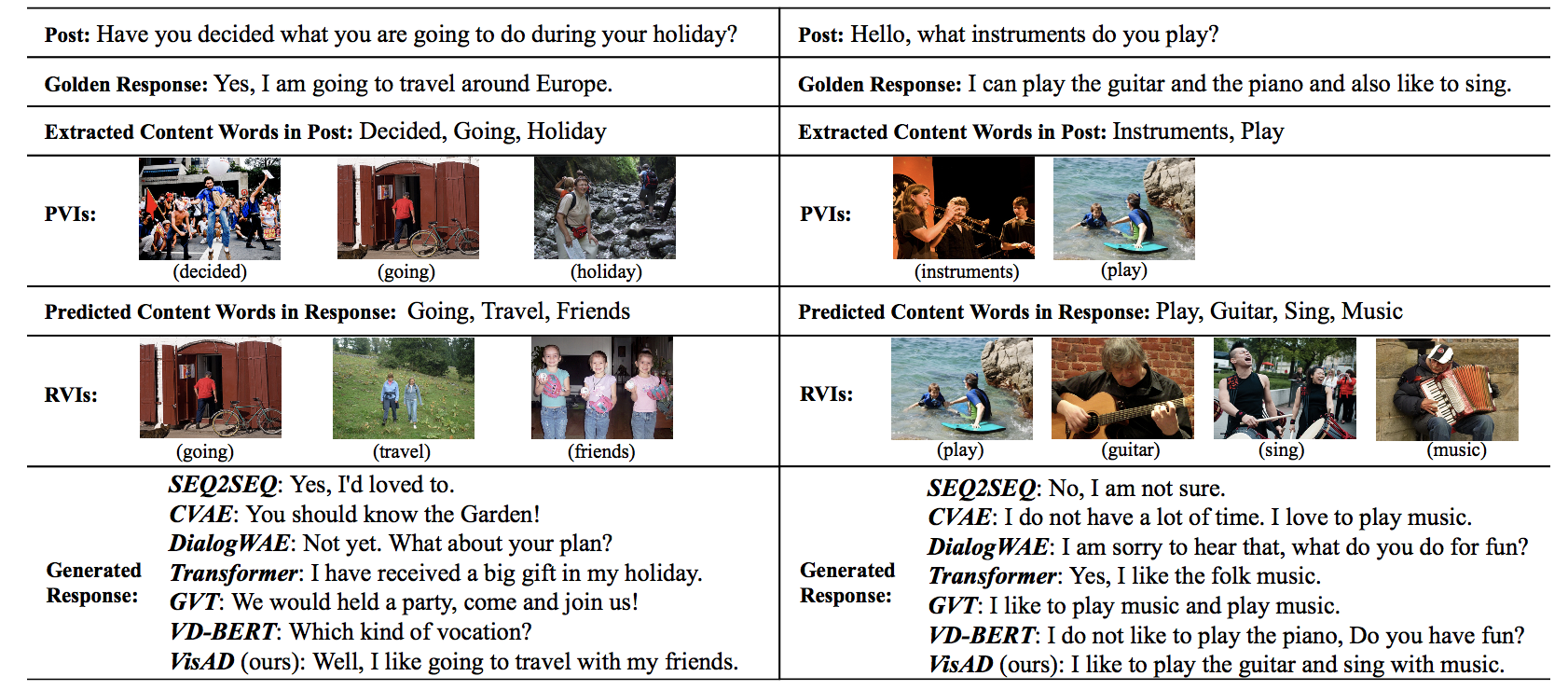}
\vspace{-3mm}
\caption{Case Study on \textit{DailyDialog} (left) and \textit{PersonaChat} (right) datasets.}
\label{fig:case_study}
\end{figure*}

\subsection{Case Study}

Figure \ref{fig:case_study} lists several responses generated by VisAD and the baseline models given two different posts. The content words in posts or responses, and corresponding PVIs or RVIs are also presented. From the left case, we notice that the content words of the post, i.e., “decided”, “going”, and “holiday”, are successfully extracted by the content word extraction process. Then, these words are used for retrieving PVIs that are in accordance with the post in a high confidence. These PVIs indicate some scenarios and objects reflected by the post and provide VisAD with visual information for better post understanding. Regarding the generated response, VisAD benefits from content word prediction and RVIs, and composes an appropriate utterance “Well, I like going to travel with my friends”. The right case in \textit{PersonaChat} can also illustrate the effectiveness of generated content words in response and RVIs for generating an informative and related response. In general, we find that VisAD is able to effectively fuse visual impressions into the pure-language dialogue response generation task.

\subsection{Further Analysis}

\paragraph{\textbf{Analysis of Word-Image Mapping Model}}
To validate the effectiveness of word-image mapping model, we present the top-5 images retrieved by content words in the left case (\textit{DailyDialog}) shown in Figure \ref{fig:case_study}. Although the word-image mapping model is trained on an image captioning dataset without word-image annotations, it still shows a strong selectivity and helps retrieve relevant images for utterances. This related visual information improves the understanding of a post and leads to a better performance on dialogue generation. Besides, we also observe that some images (marked with blue dotted boxes) are not faithfully grounded. These mis-alignments are possibly caused by the limitations of weak supervision for content words and images in our training data. In contrast, the top-1 scored images (marked with red dotted boxes), i.e., visual impressions, for extracted words are more suitable to reflect human understanding.

\begin{figure}[!t]
\centering
\includegraphics[width=0.99\linewidth]{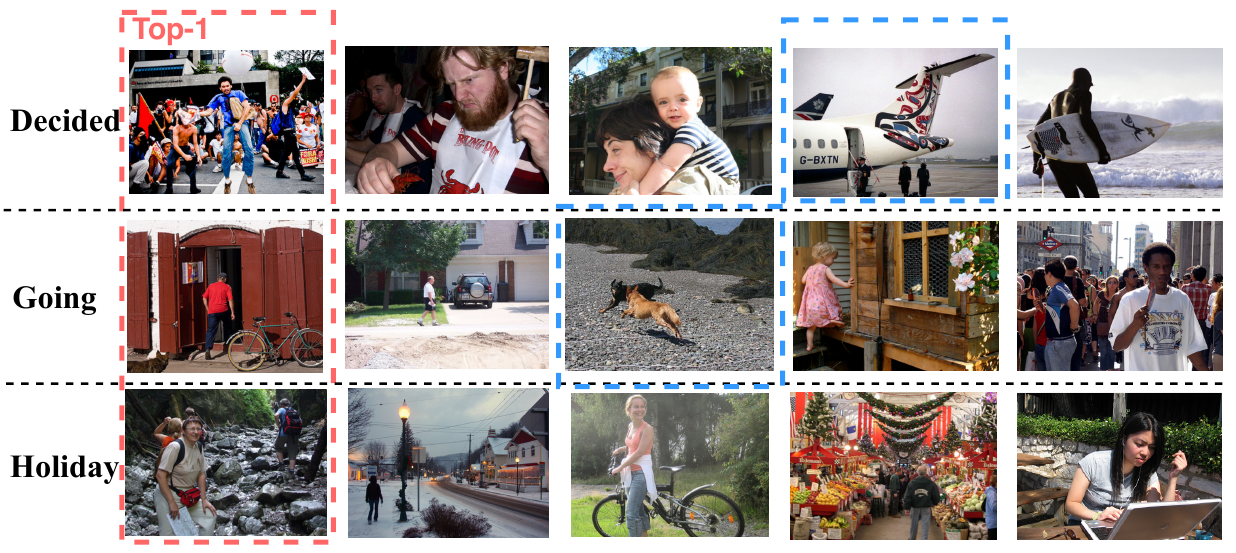}
\vspace{-3mm}
\caption{Visualization of the top-5 retrieved images for each content word from the word-image mapping model.}
\label{fig:visualiimage}
\end{figure}

\paragraph{\textbf{Analysis of Content Word Generation}} 
To validate the effectiveness of content word generation in the first sub-decoder, we devise two relevant metrics: average number of generated content words ($\#$Avg. Pred) and content word prediction accuracy (PreAcc).
The content word prediction accuracy (PreAcc) measures the relevance between the generated content words and ground-truth content words. 
The PreAcc score is defined as follows: 
\begin{equation}
\mathrm{PreAcc} = \frac{\sum_{i=1}^{N}\frac{p_i}{g_i}}{N},
\end{equation}
where $N$ is the number of samples in the test set, $g_i$ stands for the number of ground-truth content words in the $i$-th sample, and $p_i$ represents the number of content words in both generated and ground-truth sets. 
As shown in Table \ref{tab:cw}, the PreAcc scores of \textit{DailyDialog} and \textit{PersonaChat} reach 0.389 and 0.326, respectively, and there is still space for improvement. Besides, the $\#$Avg. Pred scores on two datasets are 2.66 and 2.89, respectively, which is close to the size of ground-truth content words.

\begin{table}
    \caption{Analysis on the content word generation phase. ``\#Avg. Pred'' and ``\#Avg. GT'' denote the average number of content words in generated and ground-truth sets.}
    \vspace{-1mm}
    \centering
    \begin{tabular}{l|ccc}
    \toprule[1pt] 
       Dataset  & PreAcc & \#Avg. Pred & \#Avg. GT \\ \hline
       \textit{DailyDialog} & 0.389  & 2.66 & 3.10 \\
       \textit{PersonaChat} & 0.326  & 2.89 & 3.45 \\
    \bottomrule[1pt]
    \end{tabular}
    \label{tab:cw}
\end{table}

\section{Conclusion and Future Work}

In this work, we propose a two-stage framework for open-domain dialogue generation with the assistance of visual information. Our method constructs explicit visual impressions reflecting scenarios and objects described in utterances based on human understanding. Since the post and response information do not share the same semantic space, we integrate post visual impressions (PVIs) and response visual impressions (RVIs) into the encoder and decoder for post understanding and response generation, respectively. Experimental results on two pure-language dialogue datasets show that our proposed model significantly outperforms the baselines.

For future work, we will study on the accuracy of the visual impression construction process and its influence on the second stage. Besides, since an image contains more information than a high-level abstractive word, we will try to distill more useful parts from the image for response generation process.


\section{Acknowledgements}

We would like to thank all the reviewers for their insightful and valuable comments and suggestions.



\bibliographystyle{ACM-Reference-Format}
\balance
\bibliography{acmmm2021}




\end{document}